\documentclass[sigconf]{acmart}

\AtBeginDocument{%
  \providecommand\BibTeX{{%
    \normalfont B\kern-0.5em{\scshape i\kern-0.25em b}\kern-0.8em\TeX}}}

\setcopyright{acmcopyright}
\copyrightyear{2023}
\acmYear{2023}
\setcopyright{rightsretained}
\acmConference[CHI '23 Workshop on Generative AI and HCI]{CHI 2023 Workshop on Generative AI and HCI}{Apr 28, 2023}{Virtual}
\acmBooktitle{CHI 2023 Workshop on Generative AI and HCI), Apr 28, 2023, Virtual}
\acmDOI{10.1145/3411764.3445618}

\newcommand{\ja}[1]{{\color{black} #1}}

\usepackage{multirow}
\usepackage{subcaption}
\usepackage{caption}
\usepackage{chngcntr}

\begin{document}

\title[Mapping the Design Space of Interactions in Human-AI Text Co-creation Tasks]{Mapping the Design Space of Interactions in Human-AI Text Co-creation Tasks}


\author{Zijian Ding}
\affiliation{%
  \department{College of Information Studies}
  \institution{University of Maryland, College Park}
  \country{USA}}
  
\author{Joel Chan}
\affiliation{%
  \department{College of Information Studies}
  \institution{University of Maryland, College Park}
  \country{USA}}

\renewcommand{\shortauthors}{Z. Ding \& J. Chan}

\begin{abstract}
Large Language Models (LLMs) have demonstrated impressive text generation capabilities, prompting us to reconsider the future of human-AI co-creation and how humans interact with LLMs. In this paper, we present a spectrum of content generation tasks and their corresponding human-AI interaction patterns. These tasks include: 1) fixed-scope content curation tasks with minimal human-AI interactions, 2) independent creative tasks with precise human-AI interactions, and 3) complex and interdependent creative tasks with iterative human-AI interactions. We encourage the generative AI and HCI research communities to focus on the more complex and interdependent tasks, which require greater levels of human involvement.
\end{abstract}

 


\keywords{Large Language Models, Analogy, Creativity Support Tools}


\maketitle

\begin{figure*}
\centering \includegraphics[width=0.9\textwidth]{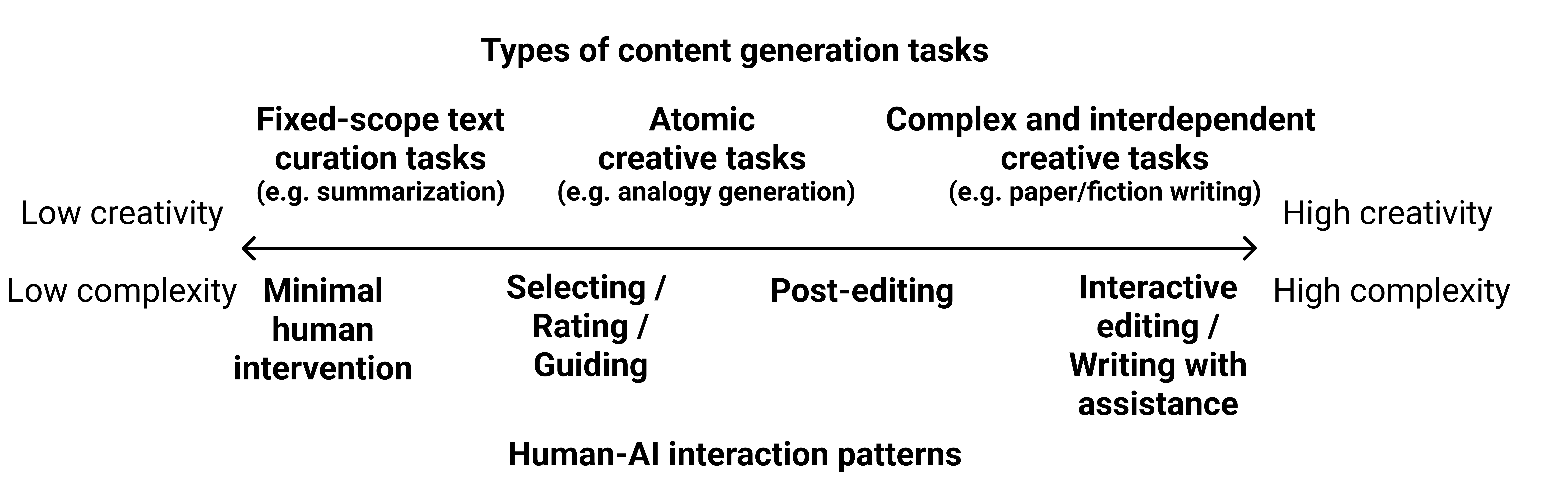}
\caption{Spectrum of human-AI co-creation tasks and corresponding human intervention complexity. The upper half describes the spectrum of text-based co-creation tasks from low to high creativity and complexity; the bottom half proposes a mapping of the points on these spectrum to human-AI interaction patterns from the taxonomy in \cite{chengMappingDesignSpace2022}.}
\label{fig:spectrum}
\Description{TODO}
\end{figure*}

\label{sec:intro}
\section{Introduction}

Large Language Models (LLMs), such as Generative Pre-trained Transformer 3 (GPT-3) \cite{brownLanguageModelsAre2020}, have garnered significant attention from researchers and practitioners for their ability to generate text content. The rapid success of ChatGPT\footnote{https://chat.openai.com/chat} - reaching 100 million monthly active users just two months after its launch and setting a record for the fastest-growing consumer application in history\footnote{https://www.reuters.com/technology/chatgpt-sets-record-fastest-growing-user-base-analyst-note-2023-02-01/} - highlights not only the potential and capabilities of Generative AI for producing precise and personalized text content, but also the critical role of interface and interaction in communicating with AI. Since ChatGPT is a variant of GPT-3 fine-tuned for conversational tasks, the technical foundation remains similar; instead, the primary difference seems to be the shift in human-AI interaction paradigms, from prompt programming, parameter tuning and autocompleting in OpenAI playground or API\footnote{https://platform.openai.com/overview}, to interactive conversations in ChatGPT. 

The wide deployment of these language models beyond academic research projects carries risks, but has also uncovered a broader sense of the potential applications of these models: from foundation models for traditional NLP applications such as text classification, summarization, and information extraction, to generative applications such as \ja{analogy generation} and even complex creative tasks such as \ja{fiction creation}. What might this wave of progress enable for augmenting --- rather than automating --- human creativity? 

To help make sense of this question, in this position paper we sketch out a possible \textbf{design space of human-AI co-creation}, focusing on the role of humans and their interactions and collaboration with Generative AI. Specifically, we synthesize previous research on human-AI interactions in text generation applications and tasks onto a spectrum of task complexity and creativity: 1) fixed-scope text curation tasks, 2) atomic creative tasks and 3) complex and interdependent creative tasks; and propose a mapping of this spectrum to a taxonomy of existing design patterns of human-AI interaction that can address the requirements and challenges of each point in the spectrum, as shown in Figure \ref{fig:spectrum}. 
Finally, we suggest future avenues for further exploring this spectrum, especially towards more complex and interdependent creative tasks.

\section{Types of human-AI interactions for text generation}

Our discussion of human-AI interaction for text generation draws on the taxonomy of five common human-AI interactions proposed by Cheng et al. \cite{chengMappingDesignSpace2022}: 1) \textit{guiding model output}, 2) \textit{selecting or rating model output}, 3) \textit{post-editing}, 4) \textit{interactive editing} (initiated by AI) and 5) \textit{writing with model assistance} (initiated by human). 

The former three interaction types do not involve rounds of iterations between human and AI, referred to as \textit{precise human-AI interactions}. In contrast, the latter two types involve rounds of iterations between human and AI, referred to as \textit{iterative human-AI interactions}. In the following sections, we discuss how these interaction types can be usefully mapped to a spectrum of text-based human-AI co-creation tasks that range from low to high complexity and creativity.

\section{Spectrum of text-based human-AI co-creation tasks}

\subsection{Minimal human-AI interactions in fixed-scope content curation tasks}

The first point in our spectrum can be described as fixed-scope content curation tasks.
Previous research has indicated that Large language models (LLMs) can effectively handle well-defined, content curation tasks such as text summarization \cite{dangTextGenerationSupporting2022, goyalNewsSummarizationEvaluation2022}, content refinement \cite{linWhyHowEmbrace2023}, and code explanation \cite{macneilExperiencesUsingCode2022, macneilGeneratingDiverseCode2022}. In these tasks, existing knowledge and information are summarized and presented in a cohesive manner, but no new knowledge is generated. In these fixed-scope tasks, advanced LLM models such as GPT-3 DaVinci have already produced satisfactory results with no human intervention on the outputs \cite{brownLanguageModelsAre2020}.
According to a study conducted by Clark et al. \cite{clarkAllThatHuman2021}, text generated by GPT-3 exhibited such a high degree of linguistic sophistication that it was almost impossible for human evaluators to discern whether it had been authored by a machine or a human, and carefully designed frameworks are required to scrutinize different types of human and machine errors and determine the authorship of a piece of text \cite{douGPT3TextIndistinguishable2022}. The aforementioned evidence serves as a testament to the high quality of machine-generated text, raising the possibility that in the future, the amount of human involvement required for content curation tasks could be significantly reduced.



\subsection{Precise human-AI interactions in atomic creative tasks}
The second point in our spectrum can be described as atomic creative tasks. By creative, we mean outputs that are both novel and useful \cite{sawyerExplainingCreativityScience2012,runcoStandardDefinitionCreativity2012}, 
including generating analogies / metaphors / analogous design concepts \cite{webbEmergentAnalogicalReasoning2022, zhuGenerativePreTrainedTransformer2022, bhavyaAnalogyGenerationPrompting2022, leeEvaluatingHumanLanguageModel2022}, slogans \cite{clarkCreativeWritingMachine2018} and inspirations for tweetorials (short technical explanations on Twitter) \cite{geroSparksInspirationScience2022}.
The broad coverage of existing knowledge by LLMs such as GPT-3 can help to create novel connections, known as "creative leaps" \cite{chanSemanticallyFarInspirations2017,siangliulueProvidingTimelyExamples2015,noyQuantitativeStudyCreative2012}. However, generating truly creative, inspiring, and insightful content often requires domain-specific knowledge, including subtle and implicit knowledge, which may not be present in the training data for LLMs. To compensate for this lack of knowledge, LLMs must be guided by carefully crafted prompts and examples, and their outputs must be selected or edited by humans to ensure their quality. In other words, these specific creative tasks demand precise human-AI interactions.

For instance, when generating analogous problems, classic analogous problems created or selected by humans, such as Duncker and Lees’ \cite{dunckerProblemsolving1945} radiation problem, can be applied to \textit{guide} LLMs in the analogy generation process. The generated analogies must then be \textit{selected} or \textit{rated}, even \textit{post-edited} by humans to avoid any biases, illegal, or inappropriate content. \ja{In some recent experiments to explore the current performance of LLMs on atomic creative tasks, we generated 120 analogous problems with GPT-3 \textit{text-davinci-002} model and the Duncker and Lees’ analogous problem for one-shot learning, and asked participants to use them to reformulate the original problem \cite{ding2023fluid}. Our results showed that the AI-generated analogous problems were frequently perceived as helpful (with a median helpfulness rating of 4 out of 5) and led to observable changes in problem formulation in approximately 80\% of cases. However, we also found that up to 25\% of the outputs were potentially harmful, mostly due to potentially upsetting content that was not biased or toxic. Our findings demonstrate the potential of using LLMs for atomic creative tasks, but also highlight the need for human intervention. Below is an example of the LLM-generated analogous problem and how participant used the analogy to stimulate reformulation of the original problem:

\begin{quote}
\textit{Original problem}

Stakeholder: owners of travel agency

Context: the restriction of pandemic has been mitigated and people are willing to travel again

Goal: reopen their traveling business

Obstacle: cannot find enough employees because people have left the travel industry during the pandemic

\textit{GPT-3 generated analogous problem}

Stakeholder: a farmer

Context: the restriction of the use of pesticides has been mitigated

Goal: use pesticides to increase crop yield

Obstacle: the farmer cannot afford to buy pesticides

\textit{Participant’s response to reformulation question}

\textbf{It's not that the travel agency can't find employees, it's that they can't afford to pay employees to work for them after being closed for so long}, thus causing a feedback loop of: not enough employees -> less money ->cant afford to hire employees -> not enough employees.
\end{quote}
}

\subsection{Iterative human-AI interactions in complex and interdependent creative tasks}

Our third and final point in the spectrum can be described as complex and interdependent creative tasks.
These are larger scale creative tasks containing a set of subtasks interdependent to each other, such as storytelling \cite{yuanWordcraftStoryWriting2022a, roemmeleWritingStoriesHelp2016, singhWhereHideStolen2022, berkovskyHowNovelistsUse2018, freiknechtProceduralGenerationInteractive2020, yangAIActiveWriter, caoLeveragingLargeLanguage2023}. Those tasks go beyond just the combination of atomic creative tasks and cannot be decomposed into atomic creative tasks. Those tasks require not only domain-specific knowledge but also the ability to plan, reason, delve into ideas, and retain context over time in order to generate new and coherent concepts, knowledge and stories. 

For these tasks, we argue that humans and large language models (LLMs) must work closely together and iteratively refine the text content, and different types of interactions are needed depending on the iteration stages. The co-creation of a story, for example, may involve a series of iterations of guidance, selection, and post-editing by both humans and LLMs \cite{yuanWordcraftStoryWriting2022a}. And fixed-scope text curation tasks and specific creative tasks can serve as building blocks for more complex and comprehensive creative tasks, such as including a literature review for scientific paper writing. Expert human review, justification, and post-editing are crucial to ensure the originality and logic of the AI-generated content and its alignment with other elements.  

\section{Future directions for text-based human-AI co-creation}
Our belief and hope is that iterative human-AI interactions in complex and interdependent creative tasks will become a focus of future research on human-AI text co-creation, due to their complexity, potential, and the need for intensive human-AI interaction. \ja{Current LLM-powered tools show potential for supporting that vision of human-AI collaboration in creative tasks, but there is still room for improvement in certain areas. For example, chatbots like ChatGPT are capable of supporting multiple rounds of interactions but currently only offer one interaction paradigm - guiding the model output. This limitation may reduce the efficiency and overall experience of creative writing with the tool.} Within the context of human-AI co-writing, Yuan et al. \cite{yuanWordcraftStoryWriting2022a} also focused primarily on exploring various formats of guiding model outputs, such as continuation, elaboration, story seeding, and infilling. While their work sheds important light on the collaborative aspects of story writing, there is potential for artificial intelligence to play an even more proactive role through interactive selecting and editing paradigms. There is also a difficult set of challenges around \textit{evaluation}. For example, NLP progress has benefited substantially from well-defined benchmarks. This can work well for accelerating progress for fixed-scope content curation tasks, but is a poor fit for atomic and complex and interdependent creative tasks with no single ``correct" reference output. Crowdsourced human evaluations may also index only surface-level linguistic coherence vs. more substantive dimensions of quality without more specific (task-specific) instructions or domain expertise: for example, Clark et al \cite{clarkAllThatHuman2021} reported that crowd workers mostly relied on form vs. content heuristics to make their judgments about human-likeness of LLM-generated text. Community-based and participatory research methods may be needed to address these challenges. A final challenge concerns the question of how to integrate domain knowledge and expertise. For example while novice users may have the most to gain from human-AI co-creation tools, they may need domain expertise to effectively control the outputs generated by AI. We believe a promising direction is to explore the design of co-creation tools that integrate the generative strengths of LLMs with sources of domain knowledge (e.g., heuristics, design patterns, knowledge bases, access to other peers and experts for feedback/validation); for example, the new Bing search tool\footnote{https://www.bing.com/new} integrates question-answering and summarization loops with API calls to knowledge bases. We believe these open problems are difficult but tractable, and look forward to exploring solutions to these problems with the human-AI interaction community.




\label{sec:example1}

\label{sec:example2}

\label{sec:example3}


\bibliographystyle{ACM-Reference-Format}
\bibliography{sample-base}


\begin{thebibliography}{29}


\ifx \showCODEN    \undefined \def \showCODEN     #1{\unskip}     \fi
\ifx \showDOI      \undefined \def \showDOI       #1{#1}\fi
\ifx \showISBNx    \undefined \def \showISBNx     #1{\unskip}     \fi
\ifx \showISBNxiii \undefined \def \showISBNxiii  #1{\unskip}     \fi
\ifx \showISSN     \undefined \def \showISSN      #1{\unskip}     \fi
\ifx \showLCCN     \undefined \def \showLCCN      #1{\unskip}     \fi
\ifx \shownote     \undefined \def \shownote      #1{#1}          \fi
\ifx \showarticletitle \undefined \def \showarticletitle #1{#1}   \fi
\ifx \showURL      \undefined \def \showURL       {\relax}        \fi
\providecommand\bibfield[2]{#2}
\providecommand\bibinfo[2]{#2}
\providecommand\natexlab[1]{#1}
\providecommand\showeprint[2][]{arXiv:#2}

\bibitem[\protect\citeauthoryear{Berkovsky, Hijikata, Rekimoto, Burnett,
  Billinghurst, and Quigley}{Berkovsky et~al\mbox{.}}{2018}]%
        {berkovskyHowNovelistsUse2018}
\bibfield{author}{\bibinfo{person}{Shlomo Berkovsky},
  \bibinfo{person}{Yoshinori Hijikata}, \bibinfo{person}{Jun Rekimoto},
  \bibinfo{person}{Margaret Burnett}, \bibinfo{person}{Mark Billinghurst},
  {and} \bibinfo{person}{Aaron Quigley}.} \bibinfo{year}{2018}\natexlab{}.
\newblock \showarticletitle{How {Novelists} {Use} {Generative} {Language}
  {Models}: {An} {Exploratory} {User} {Study}}. In
  \bibinfo{booktitle}{\emph{23rd {International} {Conference} on {Intelligent}
  {User} {Interfaces}}}. \bibinfo{publisher}{ACM}, \bibinfo{address}{Tokyo
  Japan}.
\newblock
\showISBNx{978-1-4503-4945-1}


\bibitem[\protect\citeauthoryear{Bhavya, Xiong, and Zhai}{Bhavya
  et~al\mbox{.}}{2022}]%
        {bhavyaAnalogyGenerationPrompting2022}
\bibfield{author}{\bibinfo{person}{Bhavya Bhavya}, \bibinfo{person}{Jinjun
  Xiong}, {and} \bibinfo{person}{Chengxiang Zhai}.}
  \bibinfo{year}{2022}\natexlab{}.
\newblock \bibinfo{title}{Analogy {Generation} by {Prompting} {Large}
  {Language} {Models}: {A} {Case} {Study} of {InstructGPT}}.
\newblock
\newblock
\urldef\tempurl%
\url{http://arxiv.org/abs/2210.04186}
\showURL{%
\tempurl}
\newblock
\shownote{arXiv:2210.04186 [cs].}


\bibitem[\protect\citeauthoryear{Brown, Mann, Ryder, Subbiah, Kaplan, Dhariwal,
  Neelakantan, Shyam, Sastry, Askell, Agarwal, Herbert-Voss, Krueger, Henighan,
  Child, Ramesh, Ziegler, Wu, Winter, Hesse, Chen, Sigler, Litwin, Gray, Chess,
  Clark, Berner, McCandlish, Radford, Sutskever, and Amodei}{Brown
  et~al\mbox{.}}{2020}]%
        {brownLanguageModelsAre2020}
\bibfield{author}{\bibinfo{person}{Tom~B. Brown}, \bibinfo{person}{Benjamin
  Mann}, \bibinfo{person}{Nick Ryder}, \bibinfo{person}{Melanie Subbiah},
  \bibinfo{person}{Jared Kaplan}, \bibinfo{person}{Prafulla Dhariwal},
  \bibinfo{person}{Arvind Neelakantan}, \bibinfo{person}{Pranav Shyam},
  \bibinfo{person}{Girish Sastry}, \bibinfo{person}{Amanda Askell},
  \bibinfo{person}{Sandhini Agarwal}, \bibinfo{person}{Ariel Herbert-Voss},
  \bibinfo{person}{Gretchen Krueger}, \bibinfo{person}{Tom Henighan},
  \bibinfo{person}{Rewon Child}, \bibinfo{person}{Aditya Ramesh},
  \bibinfo{person}{Daniel~M. Ziegler}, \bibinfo{person}{Jeffrey Wu},
  \bibinfo{person}{Clemens Winter}, \bibinfo{person}{Christopher Hesse},
  \bibinfo{person}{Mark Chen}, \bibinfo{person}{Eric Sigler},
  \bibinfo{person}{Mateusz Litwin}, \bibinfo{person}{Scott Gray},
  \bibinfo{person}{Benjamin Chess}, \bibinfo{person}{Jack Clark},
  \bibinfo{person}{Christopher Berner}, \bibinfo{person}{Sam McCandlish},
  \bibinfo{person}{Alec Radford}, \bibinfo{person}{Ilya Sutskever}, {and}
  \bibinfo{person}{Dario Amodei}.} \bibinfo{year}{2020}\natexlab{}.
\newblock \showarticletitle{Language {Models} are {Few}-{Shot} {Learners}}.
\newblock \bibinfo{journal}{\emph{arXiv:2005.14165 [cs]}} (\bibinfo{date}{June}
  \bibinfo{year}{2020}).
\newblock
\urldef\tempurl%
\url{http://arxiv.org/abs/2005.14165}
\showURL{%
\tempurl}
\newblock
\shownote{00030 arXiv: 2005.14165.}


\bibitem[\protect\citeauthoryear{Cao}{Cao}{2023}]%
        {caoLeveragingLargeLanguage2023}
\bibfield{author}{\bibinfo{person}{Chen Cao}.} \bibinfo{year}{2023}\natexlab{}.
\newblock \bibinfo{title}{Leveraging {Large} {Language} {Model} and
  {Story}-{Based} {Gamification} in {Intelligent} {Tutoring} {System} to
  {Scaffold} {Introductory} {Programming} {Courses}: {A} {Design}-{Based}
  {Research} {Study}}.
\newblock
\newblock
\urldef\tempurl%
\url{http://arxiv.org/abs/2302.12834}
\showURL{%
\tempurl}
\newblock
\shownote{arXiv:2302.12834 [cs].}


\bibitem[\protect\citeauthoryear{Chan, Siangliulue, Qori~McDonald, Liu,
  Moradinezhad, Aman, Solovey, Gajos, and Dow}{Chan et~al\mbox{.}}{2017}]%
        {chanSemanticallyFarInspirations2017}
\bibfield{author}{\bibinfo{person}{Joel Chan}, \bibinfo{person}{Pao
  Siangliulue}, \bibinfo{person}{Denisa Qori~McDonald}, \bibinfo{person}{Ruixue
  Liu}, \bibinfo{person}{Reza Moradinezhad}, \bibinfo{person}{Safa Aman},
  \bibinfo{person}{Erin~T. Solovey}, \bibinfo{person}{Krzysztof~Z. Gajos},
  {and} \bibinfo{person}{Steven~P. Dow}.} \bibinfo{year}{2017}\natexlab{}.
\newblock \showarticletitle{Semantically {Far} {Inspirations} {Considered}
  {Harmful}?: {Accounting} for {Cognitive} {States} in {Collaborative}
  {Ideation}}. In \bibinfo{booktitle}{\emph{Proceedings of the 2017 {ACM}
  {SIGCHI} {Conference} on {Creativity} and {Cognition}}}
  \emph{(\bibinfo{series}{C\&{C} '17})}. \bibinfo{publisher}{ACM},
  \bibinfo{address}{New York, NY, USA}, \bibinfo{pages}{93--105}.
\newblock
\showISBNx{978-1-4503-4403-6}
\urldef\tempurl%
\url{https://doi.org/10.1145/3059454.3059455}
\showDOI{\tempurl}


\bibitem[\protect\citeauthoryear{Cheng, Smith-Renner, Zhang, Tetreault, and
  Jaimes-Larrarte}{Cheng et~al\mbox{.}}{2022}]%
        {chengMappingDesignSpace2022}
\bibfield{author}{\bibinfo{person}{Ruijia Cheng}, \bibinfo{person}{Alison
  Smith-Renner}, \bibinfo{person}{Ke Zhang}, \bibinfo{person}{Joel Tetreault},
  {and} \bibinfo{person}{Alejandro Jaimes-Larrarte}.}
  \bibinfo{year}{2022}\natexlab{}.
\newblock \showarticletitle{Mapping the {Design} {Space} of {Human}-{AI}
  {Interaction} in {Text} {Summarization}}. In
  \bibinfo{booktitle}{\emph{Proceedings of the 2022 {Conference} of the {North}
  {American} {Chapter} of the {Association} for {Computational} {Linguistics}:
  {Human} {Language} {Technologies}}}. \bibinfo{publisher}{Association for
  Computational Linguistics}, \bibinfo{address}{Seattle, United States},
  \bibinfo{pages}{431--455}.
\newblock
\urldef\tempurl%
\url{https://doi.org/10.18653/v1/2022.naacl-main.33}
\showDOI{\tempurl}


\bibitem[\protect\citeauthoryear{Clark, August, Serrano, Haduong, Gururangan,
  and Smith}{Clark et~al\mbox{.}}{2021}]%
        {clarkAllThatHuman2021}
\bibfield{author}{\bibinfo{person}{Elizabeth Clark}, \bibinfo{person}{Tal
  August}, \bibinfo{person}{Sofia Serrano}, \bibinfo{person}{Nikita Haduong},
  \bibinfo{person}{Suchin Gururangan}, {and} \bibinfo{person}{Noah~A. Smith}.}
  \bibinfo{year}{2021}\natexlab{}.
\newblock \showarticletitle{All {That}'s '{Human}' {Is} {Not} {Gold}:
  {Evaluating} {Human} {Evaluation} of {Generated} {Text}}.
\newblock \bibinfo{journal}{\emph{arXiv:2107.00061 [cs]}} (\bibinfo{date}{July}
  \bibinfo{year}{2021}).
\newblock
\urldef\tempurl%
\url{http://arxiv.org/abs/2107.00061}
\showURL{%
\tempurl}
\newblock
\shownote{00008 arXiv: 2107.00061.}


\bibitem[\protect\citeauthoryear{Clark, Ross, Tan, Ji, and Smith}{Clark
  et~al\mbox{.}}{2018}]%
        {clarkCreativeWritingMachine2018}
\bibfield{author}{\bibinfo{person}{Elizabeth Clark},
  \bibinfo{person}{Anne~Spencer Ross}, \bibinfo{person}{Chenhao Tan},
  \bibinfo{person}{Yangfeng Ji}, {and} \bibinfo{person}{Noah~A. Smith}.}
  \bibinfo{year}{2018}\natexlab{}.
\newblock \showarticletitle{Creative {Writing} with a {Machine} in the {Loop}:
  {Case} {Studies} on {Slogans} and {Stories}}. In
  \bibinfo{booktitle}{\emph{23rd {International} {Conference} on {Intelligent}
  {User} {Interfaces}}}. \bibinfo{publisher}{ACM}, \bibinfo{address}{Tokyo
  Japan}, \bibinfo{pages}{329--340}.
\newblock
\showISBNx{978-1-4503-4945-1}
\urldef\tempurl%
\url{https://doi.org/10.1145/3172944.3172983}
\showDOI{\tempurl}


\bibitem[\protect\citeauthoryear{Dang, Benharrak, Lehmann, and Buschek}{Dang
  et~al\mbox{.}}{2022}]%
        {dangTextGenerationSupporting2022}
\bibfield{author}{\bibinfo{person}{Hai Dang}, \bibinfo{person}{Karim
  Benharrak}, \bibinfo{person}{Florian Lehmann}, {and} \bibinfo{person}{Daniel
  Buschek}.} \bibinfo{year}{2022}\natexlab{}.
\newblock \bibinfo{title}{Beyond {Text} {Generation}: {Supporting} {Writers}
  with {Continuous} {Automatic} {Text} {Summaries}}.
\newblock
\newblock
\urldef\tempurl%
\url{https://doi.org/10.1145/3526113.3545672}
\showDOI{\tempurl}
\newblock
\shownote{arXiv:2208.09323 [cs].}


\bibitem[\protect\citeauthoryear{Ding, Srinivasan, MacNeil, and Chan}{Ding
  et~al\mbox{.}}{2023}]%
        {ding2023fluid}
\bibfield{author}{\bibinfo{person}{Zijian Ding}, \bibinfo{person}{Arvind
  Srinivasan}, \bibinfo{person}{Stephen MacNeil}, {and} \bibinfo{person}{Joel
  Chan}.} \bibinfo{year}{2023}\natexlab{}.
\newblock \showarticletitle{Fluid Transformers and Creative Analogies:
  Exploring Large Language Models' Capacity for Augmenting Cross-Domain
  Analogical Creativity}.
\newblock \bibinfo{journal}{\emph{arXiv preprint arXiv:2302.12832}}
  (\bibinfo{year}{2023}).
\newblock


\bibitem[\protect\citeauthoryear{Dou, Forbes, Koncel-Kedziorski, Smith, and
  Choi}{Dou et~al\mbox{.}}{2022}]%
        {douGPT3TextIndistinguishable2022}
\bibfield{author}{\bibinfo{person}{Yao Dou}, \bibinfo{person}{Maxwell Forbes},
  \bibinfo{person}{Rik Koncel-Kedziorski}, \bibinfo{person}{Noah Smith}, {and}
  \bibinfo{person}{Yejin Choi}.} \bibinfo{year}{2022}\natexlab{}.
\newblock \showarticletitle{Is {GPT}-3 {Text} {Indistinguishable} from {Human}
  {Text}? {Scarecrow}: {A} {Framework} for {Scrutinizing} {Machine} {Text}}. In
  \bibinfo{booktitle}{\emph{Proceedings of the 60th {Annual} {Meeting} of the
  {Association} for {Computational} {Linguistics} ({Volume} 1: {Long}
  {Papers})}}. \bibinfo{publisher}{Association for Computational Linguistics},
  \bibinfo{address}{Dublin, Ireland}, \bibinfo{pages}{7250--7274}.
\newblock
\urldef\tempurl%
\url{https://doi.org/10.18653/v1/2022.acl-long.501}
\showDOI{\tempurl}


\bibitem[\protect\citeauthoryear{Duncker}{Duncker}{1945}]%
        {dunckerProblemsolving1945}
\bibfield{author}{\bibinfo{person}{Karl Duncker}.}
  \bibinfo{year}{1945}\natexlab{}.
\newblock \showarticletitle{On problem-solving.}
\newblock \bibinfo{journal}{\emph{Psychological Monographs}}
  \bibinfo{volume}{58}, \bibinfo{number}{5} (\bibinfo{year}{1945}),
  \bibinfo{pages}{i--113}.
\newblock
\showISSN{0096-9753}
\urldef\tempurl%
\url{https://doi.org/10.1037/h0093599}
\showDOI{\tempurl}


\bibitem[\protect\citeauthoryear{Freiknecht and Effelsberg}{Freiknecht and
  Effelsberg}{2020}]%
        {freiknechtProceduralGenerationInteractive2020}
\bibfield{author}{\bibinfo{person}{Jonas Freiknecht} {and}
  \bibinfo{person}{Wolfgang Effelsberg}.} \bibinfo{year}{2020}\natexlab{}.
\newblock \showarticletitle{Procedural {Generation} of {Interactive} {Stories}
  using {Language} {Models}}. In \bibinfo{booktitle}{\emph{International
  {Conference} on the {Foundations} of {Digital} {Games}}}.
  \bibinfo{publisher}{ACM}, \bibinfo{address}{Bugibba Malta},
  \bibinfo{pages}{1--8}.
\newblock
\showISBNx{978-1-4503-8807-8}
\urldef\tempurl%
\url{https://doi.org/10.1145/3402942.3409599}
\showDOI{\tempurl}


\bibitem[\protect\citeauthoryear{Gero, Liu, and Chilton}{Gero
  et~al\mbox{.}}{2022}]%
        {geroSparksInspirationScience2022}
\bibfield{author}{\bibinfo{person}{Katy~Ilonka Gero}, \bibinfo{person}{Vivian
  Liu}, {and} \bibinfo{person}{Lydia Chilton}.}
  \bibinfo{year}{2022}\natexlab{}.
\newblock \showarticletitle{Sparks: {Inspiration} for {Science} {Writing} using
  {Language} {Models}}. In \bibinfo{booktitle}{\emph{Designing {Interactive}
  {Systems} {Conference}}}. \bibinfo{publisher}{ACM}, \bibinfo{address}{Virtual
  Event Australia}, \bibinfo{pages}{1002--1019}.
\newblock
\showISBNx{978-1-4503-9358-4}
\urldef\tempurl%
\url{https://doi.org/10.1145/3532106.3533533}
\showDOI{\tempurl}


\bibitem[\protect\citeauthoryear{Goyal, Li, and Durrett}{Goyal
  et~al\mbox{.}}{2022}]%
        {goyalNewsSummarizationEvaluation2022}
\bibfield{author}{\bibinfo{person}{Tanya Goyal}, \bibinfo{person}{Junyi~Jessy
  Li}, {and} \bibinfo{person}{Greg Durrett}.} \bibinfo{year}{2022}\natexlab{}.
\newblock \bibinfo{title}{News {Summarization} and {Evaluation} in the {Era} of
  {GPT}-3}.
\newblock
\newblock
\urldef\tempurl%
\url{http://arxiv.org/abs/2209.12356}
\showURL{%
\tempurl}
\newblock
\shownote{arXiv:2209.12356 [cs].}


\bibitem[\protect\citeauthoryear{Lee, Srivastava, Hardy, Thickstun, Durmus,
  Paranjape, Gerard-Ursin, Li, Ladhak, Rong, Wang, Kwon, Park, Cao, Lee,
  Bommasani, Bernstein, and Liang}{Lee et~al\mbox{.}}{2022}]%
        {leeEvaluatingHumanLanguageModel2022}
\bibfield{author}{\bibinfo{person}{Mina Lee}, \bibinfo{person}{Megha
  Srivastava}, \bibinfo{person}{Amelia Hardy}, \bibinfo{person}{John
  Thickstun}, \bibinfo{person}{Esin Durmus}, \bibinfo{person}{Ashwin
  Paranjape}, \bibinfo{person}{Ines Gerard-Ursin}, \bibinfo{person}{Xiang~Lisa
  Li}, \bibinfo{person}{Faisal Ladhak}, \bibinfo{person}{Frieda Rong},
  \bibinfo{person}{Rose~E. Wang}, \bibinfo{person}{Minae Kwon},
  \bibinfo{person}{Joon~Sung Park}, \bibinfo{person}{Hancheng Cao},
  \bibinfo{person}{Tony Lee}, \bibinfo{person}{Rishi Bommasani},
  \bibinfo{person}{Michael Bernstein}, {and} \bibinfo{person}{Percy Liang}.}
  \bibinfo{year}{2022}\natexlab{}.
\newblock \bibinfo{title}{Evaluating {Human}-{Language} {Model} {Interaction}}.
\newblock
\newblock
\urldef\tempurl%
\url{http://arxiv.org/abs/2212.09746}
\showURL{%
\tempurl}
\newblock
\shownote{arXiv:2212.09746 [cs] version: 2.}


\bibitem[\protect\citeauthoryear{Lin}{Lin}{2023}]%
        {linWhyHowEmbrace2023}
\bibfield{author}{\bibinfo{person}{Zhicheng Lin}.}
  \bibinfo{year}{2023}\natexlab{}.
\newblock \bibinfo{booktitle}{\emph{Why and how to embrace {AI} such as
  {ChatGPT} in your academic life}}.
\newblock \bibinfo{type}{preprint}. \bibinfo{institution}{PsyArXiv}.
\newblock
\urldef\tempurl%
\url{https://doi.org/10.31234/osf.io/sdx3j}
\showDOI{\tempurl}


\bibitem[\protect\citeauthoryear{MacNeil, Tran, Hellas, Kim, Sarsa, Denny,
  Bernstein, and Leinonen}{MacNeil et~al\mbox{.}}{2022a}]%
        {macneilExperiencesUsingCode2022}
\bibfield{author}{\bibinfo{person}{Stephen MacNeil}, \bibinfo{person}{Andrew
  Tran}, \bibinfo{person}{Arto Hellas}, \bibinfo{person}{Joanne Kim},
  \bibinfo{person}{Sami Sarsa}, \bibinfo{person}{Paul Denny},
  \bibinfo{person}{Seth Bernstein}, {and} \bibinfo{person}{Juho Leinonen}.}
  \bibinfo{year}{2022}\natexlab{a}.
\newblock \bibinfo{title}{Experiences from {Using} {Code} {Explanations}
  {Generated} by {Large} {Language} {Models} in a {Web} {Software}
  {Development} {E}-{Book}}.
\newblock
\newblock
\urldef\tempurl%
\url{http://arxiv.org/abs/2211.02265}
\showURL{%
\tempurl}
\newblock
\shownote{arXiv:2211.02265 [cs].}


\bibitem[\protect\citeauthoryear{MacNeil, Tran, Mogil, Bernstein, Ross, and
  Huang}{MacNeil et~al\mbox{.}}{2022b}]%
        {macneilGeneratingDiverseCode2022}
\bibfield{author}{\bibinfo{person}{Stephen MacNeil}, \bibinfo{person}{Andrew
  Tran}, \bibinfo{person}{Dan Mogil}, \bibinfo{person}{Seth Bernstein},
  \bibinfo{person}{Erin Ross}, {and} \bibinfo{person}{Ziheng Huang}.}
  \bibinfo{year}{2022}\natexlab{b}.
\newblock \showarticletitle{Generating {Diverse} {Code} {Explanations} using
  the {GPT}-3 {Large} {Language} {Model}}. In
  \bibinfo{booktitle}{\emph{Proceedings of the 2022 {ACM} {Conference} on
  {International} {Computing} {Education} {Research} - {Volume} 2}}.
  \bibinfo{publisher}{ACM}, \bibinfo{address}{Lugano and Virtual Event
  Switzerland}, \bibinfo{pages}{37--39}.
\newblock
\showISBNx{978-1-4503-9195-5}
\urldef\tempurl%
\url{https://doi.org/10.1145/3501709.3544280}
\showDOI{\tempurl}


\bibitem[\protect\citeauthoryear{Noy, Hart, Andrew, Ramote, Mayo, and Alon}{Noy
  et~al\mbox{.}}{2012}]%
        {noyQuantitativeStudyCreative2012}
\bibfield{author}{\bibinfo{person}{Lior Noy}, \bibinfo{person}{Yuval Hart},
  \bibinfo{person}{Natalie Andrew}, \bibinfo{person}{Omer Ramote},
  \bibinfo{person}{Avi Mayo}, {and} \bibinfo{person}{Uri Alon}.}
  \bibinfo{year}{2012}\natexlab{}.
\newblock \showarticletitle{A quantitative study of creative leaps},
  \bibfield{editor}{\bibinfo{person}{Mary Maher}, \bibinfo{person}{Kristian
  Hammond}, \bibinfo{person}{Alison Pease},
  \bibinfo{person}{{\{\vphantom{\}}Rafael Pérez}}, \bibinfo{person}{Dan
  Ventura}, {and} \bibinfo{person}{Geraint Wiggins}} (Eds.).
  \bibinfo{pages}{72--76}.
\newblock
\urldef\tempurl%
\url{http://computationalcreativity.net/iccc2012/wp-content/uploads/2012/05/072-Noy.pdf}
\showURL{%
\tempurl}


\bibitem[\protect\citeauthoryear{Roemmele}{Roemmele}{2016}]%
        {roemmeleWritingStoriesHelp2016}
\bibfield{author}{\bibinfo{person}{Melissa Roemmele}.}
  \bibinfo{year}{2016}\natexlab{}.
\newblock \showarticletitle{Writing {Stories} with {Help} from {Recurrent}
  {Neural} {Networks}}.
\newblock \bibinfo{journal}{\emph{Proceedings of the AAAI Conference on
  Artificial Intelligence}} \bibinfo{volume}{30}, \bibinfo{number}{1}
  (\bibinfo{date}{March} \bibinfo{year}{2016}).
\newblock
\showISSN{2374-3468, 2159-5399}
\urldef\tempurl%
\url{https://doi.org/10.1609/aaai.v30i1.9810}
\showDOI{\tempurl}


\bibitem[\protect\citeauthoryear{Runco and Jaeger}{Runco and Jaeger}{2012}]%
        {runcoStandardDefinitionCreativity2012}
\bibfield{author}{\bibinfo{person}{Mark~A. Runco} {and}
  \bibinfo{person}{Garrett~J. Jaeger}.} \bibinfo{year}{2012}\natexlab{}.
\newblock \showarticletitle{The {Standard} {Definition} of {Creativity}}.
\newblock \bibinfo{journal}{\emph{Creativity Research Journal}}
  \bibinfo{volume}{24}, \bibinfo{number}{1} (\bibinfo{date}{Jan.}
  \bibinfo{year}{2012}), \bibinfo{pages}{92--96}.
\newblock
\showISSN{1040-0419}
\urldef\tempurl%
\url{https://doi.org/10.1080/10400419.2012.650092}
\showDOI{\tempurl}


\bibitem[\protect\citeauthoryear{Sawyer}{Sawyer}{2012}]%
        {sawyerExplainingCreativityScience2012}
\bibfield{author}{\bibinfo{person}{R.~Keith Sawyer}.}
  \bibinfo{year}{2012}\natexlab{}.
\newblock \bibinfo{booktitle}{\emph{Explaining creativity: the science of human
  innovation} (\bibinfo{edition}{2nd} ed.)}.
\newblock \bibinfo{publisher}{Oxford University Press}, \bibinfo{address}{New
  York}.
\newblock
\showISBNx{0-19-973757-6}


\bibitem[\protect\citeauthoryear{Siangliulue, Chan, Gajos, and Dow}{Siangliulue
  et~al\mbox{.}}{2015}]%
        {siangliulueProvidingTimelyExamples2015}
\bibfield{author}{\bibinfo{person}{Pao Siangliulue}, \bibinfo{person}{Joel
  Chan}, \bibinfo{person}{Krzysztof Gajos}, {and} \bibinfo{person}{Steven~P.
  Dow}.} \bibinfo{year}{2015}\natexlab{}.
\newblock \showarticletitle{Providing timely examples improves the quantity and
  quality of generated ideas}. In \bibinfo{booktitle}{\emph{Proceedings of the
  {ACM} {Conference} on {Creativity} and {Cognition}}}.
\newblock
\urldef\tempurl%
\url{https://doi.org/10.1145/2757226.2757230}
\showDOI{\tempurl}


\bibitem[\protect\citeauthoryear{Singh, Bernal, Savchenko, and Glassman}{Singh
  et~al\mbox{.}}{2022}]%
        {singhWhereHideStolen2022}
\bibfield{author}{\bibinfo{person}{Nikhil Singh}, \bibinfo{person}{Guillermo
  Bernal}, \bibinfo{person}{Daria Savchenko}, {and} \bibinfo{person}{Elena~L.
  Glassman}.} \bibinfo{year}{2022}\natexlab{}.
\newblock \showarticletitle{Where to {Hide} a {Stolen} {Elephant}: {Leaps} in
  {Creative} {Writing} with {Multimodal} {Machine} {Intelligence}}.
\newblock \bibinfo{journal}{\emph{ACM Transactions on Computer-Human
  Interaction}} (\bibinfo{date}{Feb.} \bibinfo{year}{2022}),
  \bibinfo{pages}{3511599}.
\newblock
\showISSN{1073-0516, 1557-7325}
\urldef\tempurl%
\url{https://doi.org/10.1145/3511599}
\showDOI{\tempurl}


\bibitem[\protect\citeauthoryear{Webb, Holyoak, and Lu}{Webb
  et~al\mbox{.}}{2022}]%
        {webbEmergentAnalogicalReasoning2022}
\bibfield{author}{\bibinfo{person}{Taylor Webb}, \bibinfo{person}{Keith~J.
  Holyoak}, {and} \bibinfo{person}{Hongjing Lu}.}
  \bibinfo{year}{2022}\natexlab{}.
\newblock \bibinfo{title}{Emergent {Analogical} {Reasoning} in {Large}
  {Language} {Models}}.
\newblock
\newblock
\urldef\tempurl%
\url{http://arxiv.org/abs/2212.09196}
\showURL{%
\tempurl}
\newblock
\shownote{arXiv:2212.09196 [cs].}


\bibitem[\protect\citeauthoryear{Yang, Zhou, Zhang, and Li}{Yang
  et~al\mbox{.}}{[n.d.]}]%
        {yangAIActiveWriter}
\bibfield{author}{\bibinfo{person}{Daijin Yang}, \bibinfo{person}{Yanpeng
  Zhou}, \bibinfo{person}{Zhiyuan Zhang}, {and} \bibinfo{person}{Toby Jia-Jun
  Li}.} \bibinfo{year}{[n.d.]}\natexlab{}.
\newblock \showarticletitle{{AI} as an {Active} {Writer}: {Interaction}
  strategies with generated text in human-{AI} collaborative fiction writing}.
\newblock  (\bibinfo{year}{[n.\,d.]}).
\newblock


\bibitem[\protect\citeauthoryear{Yuan, Coenen, Reif, and Ippolito}{Yuan
  et~al\mbox{.}}{2022}]%
        {yuanWordcraftStoryWriting2022a}
\bibfield{author}{\bibinfo{person}{Ann Yuan}, \bibinfo{person}{Andy Coenen},
  \bibinfo{person}{Emily Reif}, {and} \bibinfo{person}{Daphne Ippolito}.}
  \bibinfo{year}{2022}\natexlab{}.
\newblock \showarticletitle{Wordcraft: {Story} {Writing} {With} {Large}
  {Language} {Models}}. In \bibinfo{booktitle}{\emph{27th {International}
  {Conference} on {Intelligent} {User} {Interfaces}}}.
  \bibinfo{publisher}{ACM}, \bibinfo{address}{Helsinki Finland},
  \bibinfo{pages}{841--852}.
\newblock
\showISBNx{978-1-4503-9144-3}
\urldef\tempurl%
\url{https://doi.org/10.1145/3490099.3511105}
\showDOI{\tempurl}


\bibitem[\protect\citeauthoryear{Zhu and Luo}{Zhu and Luo}{2022}]%
        {zhuGenerativePreTrainedTransformer2022}
\bibfield{author}{\bibinfo{person}{Q. Zhu} {and} \bibinfo{person}{J. Luo}.}
  \bibinfo{year}{2022}\natexlab{}.
\newblock \showarticletitle{Generative {Pre}-{Trained} {Transformer} for
  {Design} {Concept} {Generation}: {An} {Exploration}}.
\newblock \bibinfo{journal}{\emph{Proceedings of the Design Society}}
  \bibinfo{volume}{2} (\bibinfo{date}{May} \bibinfo{year}{2022}),
  \bibinfo{pages}{1825--1834}.
\newblock
\showISSN{2732-527X}
\urldef\tempurl%
\url{https://doi.org/10.1017/pds.2022.185}
\showDOI{\tempurl}
\newblock
\shownote{Publisher: Cambridge University Press.}


\end{thebibliography}

\end{document}